\documentclass{article}

\usepackage[latin1]{inputenc}
\usepackage[T1]{fontenc}
\usepackage{pstricks,xcolor}
\usepackage{pst-tree}
\usepackage{xspace,url}
\usepackage{amsmath, amssymb}
\usepackage{hyperref}
\usepackage{ijcai07}
\numberwithin{equation}{section}

\newcommand{\ie}{i.e.\xspace}
\newcommand{\eg}{e.g.\xspace}

\newcommand{\monacro}[1]{\textsc{#1}\xspace}
\newcommand{\xml}{\monacro{xml}} 

\newcommand{\pta}{\monacro{pta}}
\newcommand{\wta}{\monacro{wta}}
\newcommand{\wha}{\monacro{wha}}
\newcommand{\wsta}{\monacro{wsta}}

\newcommand{\hmms}{\textsc{hmm}s\xspace}

\newcommand{\ptas}{\textsc{pta}\xspace}

\newcommand{\setfont}{\mathbb}

\newcommand{\R}{\ensuremath{\setfont{R}}\xspace}
\newcommand{\Q}{\ensuremath{\setfont{Q}}\xspace}

\def\argmax{\operatornamewithlimits{arg\,max}}

\title{On Probability Distributions for Trees: Representations,
  Inference and Learning}

\author{
 François Denis \and Amaury Habrard \\
   Laboratoire d'Informatique Fondamentale de Marseille (L.I.F.) \\
   UMR CNRS 6166 --- \texttt{http://www.lif.univ-mrs.fr}
   \AND Rémi Gilleron \and Marc Tommasi \\
   INRIA Futurs and Lille University, LIFL, Mostrare Project \\
   \texttt{http://www.grappa.univ-lille3.fr/mostrare}
   \AND Édouard Gilbert \\
   ÉNS de Cachan, Brittany extension\\
   INRIA Futurs and Lille University, LIFL, Mostrare Project
}

\begin{document}

\maketitle

\vspace*{2.3cm}

We study probability distributions over free algebras of
trees. Probability distributions can be seen as particular
\emph{(formal power) tree
  series}~\cite{BerstelReutenauer82,EsikKuich03}, i.e. mappings from
trees to a semiring \( K \). A widely studied class of tree
series is the class of \emph{rational} (or \emph{recognizable}) tree series
which can be defined either in an algebraic way or by means of
multiplicity tree automata. We argue that the algebraic
representation is very convenient to model probability distributions
over a free algebra of trees. First, as in the string case, the
algebraic representation allows to design learning algorithms for the
whole class of probability distributions defined by rational tree
series. Note that learning algorithms for rational tree series
correspond to learning algorithms for weighted tree automata where
both the structure and the weights are learned. 
Second, the algebraic
representation can be easily extended to deal with unranked trees
(like \xml trees where a symbol may have an unbounded number of
children).  Both properties are particularly relevant for
applications: nondeterministic automata are required for the
inference problem to be relevant (recall that Hidden Markov Models are
equivalent to nondeterministic string automata); nowadays
applications for Web Information Extraction, Web Services and document
processing consider unranked trees.

\section{Representation Issues}

Trees, either ranked or unranked, arise in many application domains to
model data. For instance \xml documents are unranked trees; in natural
language processing (NLP), syntactic structure can often be considered
as treelike. From a machine learning perspective, dealing with tree
structured data often requires to design probability distributions
over sets of trees. This problem has been addressed mainly in the NLP
community with tools like probabilistic context free grammars
\cite{ManningSchutze99}.

\newpage
\vspace*{2.3cm}

Weighted tree automata and tree series are powerful
tools to deal with tree structured data. In particular,
probabilistic tree automata and stochastic series, which both define
probability distributions on trees, allow to generalize usual
techniques from probabilistic word automata (or hidden markov models)
and series.

\paragraph{Tree Series and Weighted Tree Automata}

In these first two paragraphs, we only consider the case of ranked trees.
A tree series is a mapping from the set of trees into some semiring $K$.
Motivated by defining probability distributions, we mainly consider
the case $K=\R$. A \emph{recognizable tree series}~\cite{BerstelReutenauer82}
$S$ is defined
by a finite dimensional vector space $V$ over $K$, a mapping $\mu$
which maps every symbol of arity $p$ into a multilinear mapping
from $V^p$ into $V$ ($\mu$ uniquely extends into a morphism
from the set of trees into $V$), and a linear form $\lambda$.
$S(t)$ is defined to be $\lambda(\mu(t))$. Tree series can also be defined
by \emph{weighted tree automata} (\wta). A \wta $A$ is a tree automaton
in which every rule is given a weight in $K$. For every run $r$
on a tree $t$ (computation of the automaton according to rules over $t$),
a weight $A(t,r)$ is computed multiplying weights of rules used in the run
and the final weight of the state at the root of the tree. The weight
$A(t)$ is the sum of all $A(t,r)$ for all runs $r$ over $t$.

For commutative semirings, recognizable tree series in the algebraic
sense and in the automata sense coincide because there is an
equivalence between summation at every step and summation over all
runs. It can be shown, as in the string case, that the set of
recognizable tree series defined by deterministic \wta is strictly
included in the set of recognizable tree series. A Myhill-Nerode
Theorem can be defined for \wta over fields~\cite{Borchardt03}. 

\paragraph{Probability Distributions and Probabilistic Tree Automata}

A probability distribution $S$ over trees is a tree series such that,
for every $t$, $S(t)$ is between 0 and 1, and such that the sum of all
$S(t)$ is equal to 1. Probabilistic tree automata (\pta) are \wta
verifying normalization conditions over weights of rules and weights
of final states. They extend probabilistic automata for strings
and we recall that nondeterministic probabilistic string automata are
equivalent to hidden Markov models (\hmms). As in the string case~%
\cite{DenisEspositoHabrard06}, not
all probability distributions defined by \wta can be defined by
\pta. However, we have proved that any distribution defined by a \wta\/
\emph{with non-negative coefficients} can defined by a \pta, too.

While in the string case, every probabilistic automaton defines
a probability distribution, this is no longer true in the tree
case. Similarly to probabilistic context-free
grammars~\cite{Wetherell80}, probabilistic automata may define
inconsistent (or improper) probability distributions: the probability
of all trees is less than one. We have defined a sufficient condition
for a \pta to define a probability distribution and a polynomial time
algorithm for checking this condition.


\paragraph{Towards unranked trees}

Until this point, we only have considered ranked trees. However, unranked
trees can be expressed by ranked ones using an isomorphism defined by an
algebraic formulation~(\cite{tata97}, chapter~8). It consists in using
the right adjonction operator defined by
\( f(t_1, \dotsc, t_{n-1}) @ t_n = f(t_1, \dotsc, t_n) \); any tree
can then be written as an expression whose only operator is \( @ \), and
thus as a binary tree: \eg, $b(a,a,c(a,a))$ corresponds to
$@(@(@(b,a),a),@(@(c,a),a))$. \wta for
unranked trees can be defined as \wta for ranked trees applied to the
algebraic formulation. We call such automata weighted stepwise tree
automata (\wsta).

Hedge automata are automata for unranked trees. Each rule of a hedge
automaton~\cite{tata97} is written \( f(L) \rightarrow q \) where \( L \)
is a regular language of word with the set of states of the automata as
its alphabet. For weighted hedge automata (\wha), the weight of the
rule \( f(u) \rightarrow q \) is the product of a weight given to
the whole rule \( f(L) \rightarrow q \) and the weight of \( u \)
according to a weighted word automata associated to \( f(L) \rightarrow q \).
 When \( K \) is commutative, \wsta and \wha
define the same weight distributions on unranked trees.

Probabilistic hedge automata can be defined by adding the same kind of
summation conditions than on \wha,
but it has yet to be shown that they can be expressed by \pta through
algebraic formulation. We don't know yet weither defining series on unranked
trees directly is possible, although it can be achieved using the algebraic
formulation.

\section{Learning Probability Distributions}

\paragraph{Inference and Training}
\pta can be considered as generative models for trees. The two
classical \textbf{inference problems} are : given a \pta $A$ and given
a tree $t$, compute $p(t)$ which is defined to the sum over all of all
$p(t,r)$; and given a tree $t$, find the most likely (or Viterbi)
labeling (run) $\hat{r}$ for $t$, \ie compute $\hat{r} = \argmax_{r}
p(r | t)$.  It should be noted that the inference problems are
relevant only for nondeterministic \ptas. The \textbf{training
  problem} is: given a sample set $S$ of trees and a \pta, learn the
best real-valued parameter vector (weights assigned to rules and to
states) according to some criteria. For instance, the likelihood of
the sample set or the likelihood of the sample over Viterbi
derivations. Classical algorithms for inference (the message passing
algorithm) and learning (the Baum-Welch algorithm) can be designed for
\pta over ranked trees and unranked trees.

\paragraph{Learning Weighted Automata}

The \textbf{learning problem} extends over the training
problem. Indeed, for the training problem, the structure of the \pta
is given by the set of rules and only weights have to be found. In the
learning problem, the structure of the target automaton is
unknown. The learning problem is: given a sample set $S$ of trees drawn
according to a target rational probability distribution, learn a \wta
according to some criteria. If the probability distribution is defined
by a deterministic \pta, a learning algorithm extending over the
unweighted case has been defined
in~\cite{CarrascoOncinaCalera01}. However, this algorithm works only
for deterministic \pta. We recall that the class of probability
distributions defined by deterministic \pta is strictly included in
the class of probability distributions defined by \pta~\cite{Borchardt03}.

\paragraph{Learning Recognizable Tree Series}


and thus learning \wta can be achieved thanks to an algorithm proposed
by Denis and Habrard~\cite{DenisHabrard07}. This algorithm, which benefits
from the existence of a canonical linear representation of series, can be
applied to series which take their values in \R or \Q to learn stochastic
tree languages. It should be noted that
the algebraic view allows to learn probability distributions defined
by nondeterministic \wta. Learning probability distributions for
unranked trees is ongoing work.


\bibliography{../Bibtex/bibapp}

\end{document}